\documentclass{article} % For LaTeX2e
\usepackage[preprint]{colm2026_conference}
\usepackage{microtype}
\usepackage{hyperref}
\usepackage{url}
\usepackage{booktabs}
\usepackage{multirow}
\usepackage{placeins}
\usepackage{amsmath}
\usepackage[justification=raggedright,singlelinecheck=false]{caption}
\usepackage{lineno}
\usepackage{graphicx}
\newlength{\titleiconwidth}
\definecolor{darkblue}{rgb}{0, 0, 0.5}
\hypersetup{colorlinks=true, citecolor=darkblue, linkcolor=darkblue, urlcolor=darkblue}

\usepackage[absolute,overlay]{textpos}
\usepackage{graphicx}

\newcommand{\placechronosicon}{%
  \begin{textblock*}{20mm}(25mm,26.5mm) % (x,y)
    \includegraphics[height=13mm]{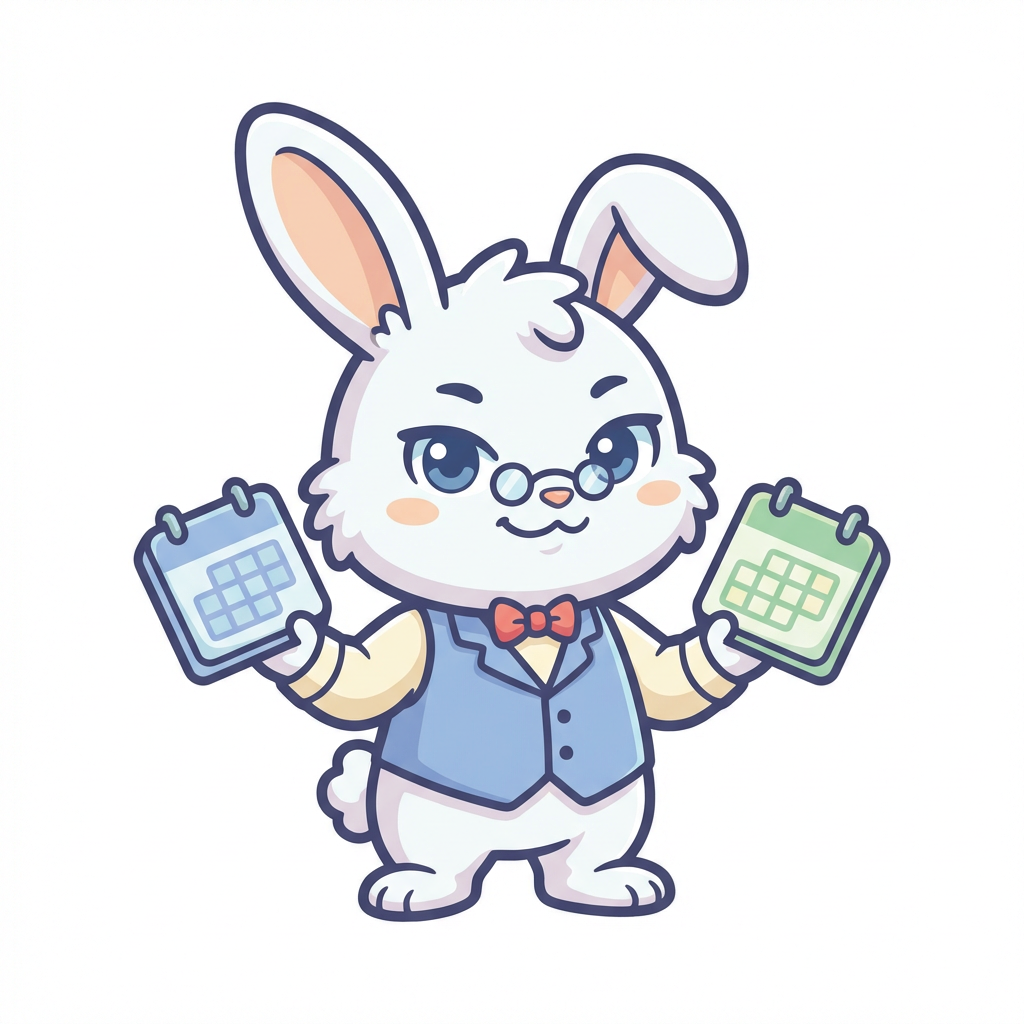}
  \end{textblock*}
}

\title{Chronos: Temporal-Aware Conversational Agents with \\Structured Event Retrieval for Long-Term Memory}

\author{Sahil Sen \\
Commercial Technology and Innovation Office, \\
PricewaterhouseCoopers \\
U.S. \\
\texttt{sahil.s.sen@pwc.com}
\And
Elias Lumer \\
Commercial Technology and Innovation Office, \\
PricewaterhouseCoopers \\
U.S. \\
\texttt{elias.lumer@pwc.com}
\And
Anmol Gulati \\
Commercial Technology and Innovation Office, \\
PricewaterhouseCoopers \\
U.S. \\
\And
Vamse Kumar Subbiah  \\
Commercial Technology and Innovation Office, \\
PricewaterhouseCoopers \\
U.S. \\
}

\begin{document}
\placechronosicon

%\ifcolmsubmission
%\linenumbers
%\fi

\maketitle

\begin{abstract}
Recent advances in Large Language Models (LLMs) have enabled conversational
AI agents to engage in extended multi-turn interactions spanning weeks or
months. However, existing memory systems struggle to reason over temporally
grounded facts and preferences that evolve across months of interaction and
lack effective retrieval strategies for multi-hop, time-sensitive queries
over long dialogue histories. We introduce Chronos, a novel temporal-aware memory
framework that decomposes raw dialogue into subject-verb-object event tuples
with resolved datetime ranges and entity aliases, indexing them in a
structured event calendar alongside a turn calendar that preserves full
conversational context. At query time, Chronos applies dynamic prompting to
generate tailored retrieval guidance for each question, directing the agent
on what to retrieve, how to filter across time ranges, and how to approach
multi-hop reasoning through an iterative tool-calling loop over both
calendars. We evaluate Chronos with 8 LLMs, both open-source and
closed-source, on the LongMemEvalS benchmark comprising 500 questions
spanning six categories of dialogue history tasks. Chronos Low achieves 92.60\% and Chronos High scores 95.60\% accuracy, setting a new
state of the art with an improvement of 7.67\% over the best prior system. Ablation results reveal the events calendar accounts for a 58.9\% gain on the baseline while all other components yield improvements between 15.5\% and 22.3\%.
Notably, Chronos Low alone surpasses prior approaches evaluated
under their strongest model configurations.
\end{abstract}
\begin{figure}[t]
    \centering
    \includegraphics[width=0.8\linewidth, clip]{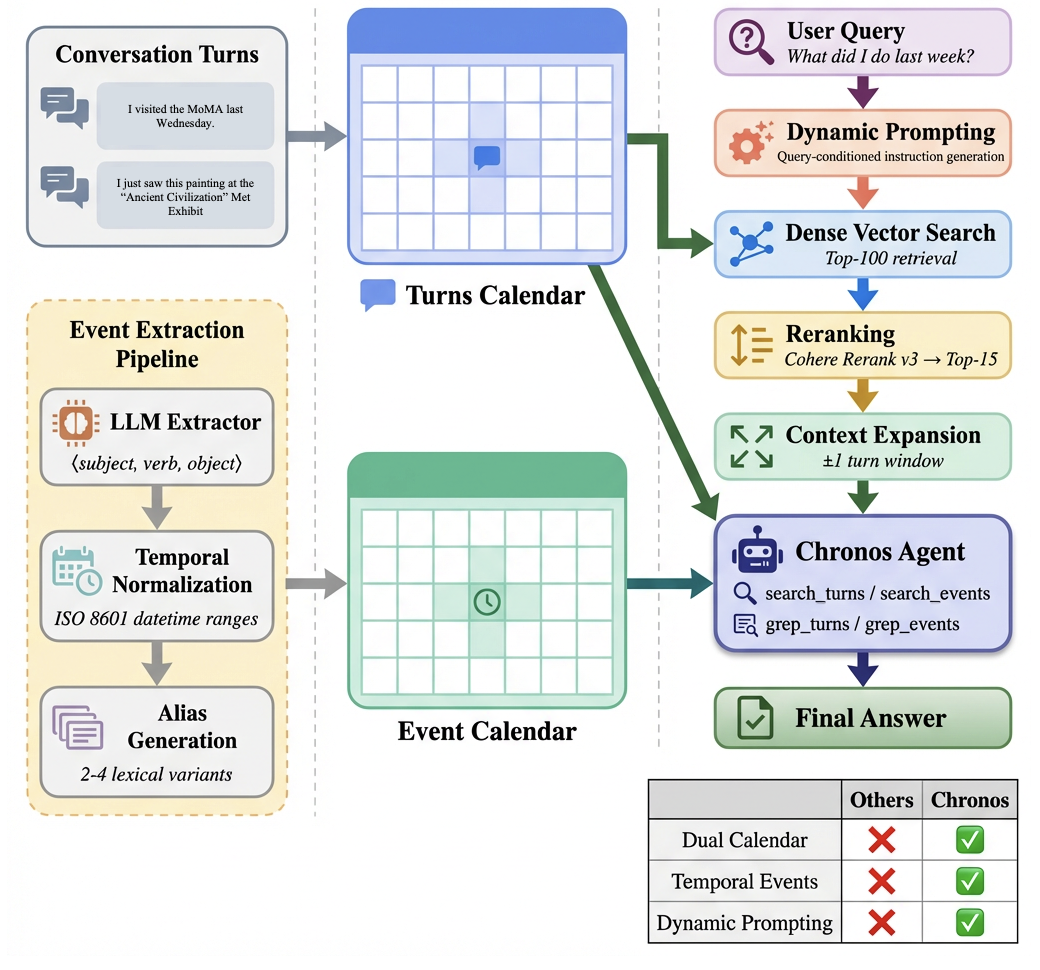}
    \caption{An overview of the Chronos Architecture. Event Extraction, Dual Indexing, and Query Processing result in a generated answer.}
    \label{fig:final_output}
\end{figure}

\section{Introduction}

The rapid progress of Large Language Models (LLMs) has enabled conversational AI agents to maintain contextual awareness across extended multi-turn interactions, supporting personalized assistance over weeks or months of conversation history \cite{llm_memory}. With breakthroughs in RAG for conversational memory, LLM agents can efficiently access historical information without exhausting context window limits (\cite{rag_survey}). As these systems are deployed in domains requiring persistent user engagement, the ability to accurately recall, track, and reason over temporally grounded events across sessions becomes essential.

Despite these advancements, conversational memory systems have struggled to find the right balance between structured knowledge building and retrieval simplicity. Systems employing comprehensive knowledge graphs extract all facts and relationships at ingestion time, building elaborate graph structures with entity resolution, fact validation, and temporal metadata (\cite{zep}). However, this global extraction creates large knowledge bases even when queries require only a subset of information. Simpler turn-level retrieval approaches avoid this overhead through direct dense-sparse hybrid search over conversation turns, but lack the structured temporal representations needed for time-sensitive queries involving date calculations or cross-session event aggregation (\cite{emergence}). Recent systems have introduced background reasoning pipelines that generate derived facts, timelines, and behavioral patterns through offline "dreaming" or observational analysis, but these query-independent deductions introduce context entropy when the precomputed knowledge proves irrelevant to specific questions (\cite{honcho}). While these systems employ LLM-based temporal normalization, they rely on global extraction strategies that process all conversational content uniformly, and they only normalize time within fact strings. The core challenge remains: comprehensive memory building introduces overhead and context entropy through over-structuring, while pure turn-level retrieval lacks temporal grounding for time-sensitive reasoning. No existing approach achieves query-conditioned selective extraction, structuring only the temporal information relevant to answering specific questions while preserving conversational context for semantic understanding.

In this paper, we introduce Chronos (visualized in Figure~\ref{fig:final_output}), a conversational memory framework centered on query-conditioned selective extraction. Chronos outperforms both pure turn-level retrieval and comprehensive knowledge-base construction. Rather than extracting all facts and relationships during ingestion or generating derived knowledge through background reasoning pipelines, Chronos performs targeted event extraction focused on temporally-grounded state transitions and timestamped occurrences, indexed alongside raw conversation turns. This approach extracts only what is necessary while avoiding the context entropy introduced by comprehensive fact extraction or query-independent background deductions. To our knowledge, Chronos is the first architecture that combines the simplicity of turn-level retrieval with selective temporal event extraction: conversation turns provide full conversational context for semantic understanding, while extracted events with structured datetime ranges enable precise temporal filtering and cross-session aggregation. In addition, Chronos implements dynamic prompting, extending query rewriting from the RAG literature to long-term memory. Rather than reformulating the search query, Chronos analyzes each question and generates tailored retrieval guidance for the agent. By structuring exactly what LLMs struggle with (temporal deltas, event sequences, date calculations) and leaving the rest as natural language turns, Chronos achieves the minimal sufficient abstraction for conversational memory.

We evaluate Chronos on the LongMemEvalS benchmark \cite{longmemeval} comprising 500 questions across six categories. Chronos Low achieves 92.60\% accuracy, establishing state-of-the-art performance (+7.67\% relative to EmergenceMem Internal). Chronos High achieves 95.60\% accuracy, the highest reported performance on this benchmark (+3.02\% relative to Mastra's OM).

\section{Related Work}\label{sec:related_work}

We organize existing work in conversational memory along four themes---the distinction between short-term and long-term memory, knowledge accumulation strategies, retrieval architectures, and summarization and fact extraction---highlighting how each reveals a gap in temporal structuring that Chronos addresses.

\subsection{Long-Term Conversational Memory}\label{subsec:longterm_memory}

Modern LLM context windows largely address short-term, or within-session, memory, allowing models to attend to prior turns without explicit memory mechanisms \cite{llm_memory}. Long-term, or across-session, memory presents a harder challenge: retaining and retrieving information from conversations that occurred days or months earlier and no longer reside in context. The model has to keep track of relationships over long periods of time, be able to connect and aggregate discrete events in the user's history, and understand changes to user preferences. There are two primary benchmarks that evaluate a model's long-term memory capabilities. LongMemEvalSevaluates 500 questions across six categories, including knowledge-update tracking, multi-session aggregation, and temporal reasoning (\cite{longmemeval}). LoCoMo evaluates memory over naturalistic human conversations spanning up to 35 sessions, measuring single-hop, multi-hop, temporal, and adversarial question answering (\cite{locomo}). However, prior work has noted several limitations: most sessions fit within modern context windows, the dataset does not evaluate knowledge updates (a key part of long-term memory), and a variety of technical errors in the benchmark itself (\cite{locomo_bad}). Notably, neither benchmark isolates the role of temporal structuring in long-term memory, leaving open the question of how much explicit temporal representation is needed for accurate cross-session reasoning.

\subsection{Knowledge Accumulation and Representations}\label{subsec:knowledge_representation}

Conversational memory systems must decide how to represent the knowledge they extract from dialogue. Broadly, representations fall along a spectrum from structured to unstructured. At the structured end, knowledge graphs encode entities and relationships as typed edges, enabling explicit reasoning over connections. Graph-based systems allow information to be retrieved logically and quickly. Simpler knowledge-graph architectures rely on a two-stage approach, first generating the structure and then filling in the details (\cite{local_to_global}). More complex systems integrate three or more different types of nodes (\cite{zep}). On the unstructured end, many systems preserve raw conversational text or generate natural language summaries and observations. Some approaches deploy specialized agents that monitor conversations and produce timestamped annotations (\cite{mastra}). Others organize memory into hierarchical tiers that progressively consolidate episodic details into semantic summaries (\cite{memgpt}). Hybrid approaches maintain both raw dialogue and extracted atomic facts as parallel corpora, preserving original context alongside structured knowledge \cite{honcho, emergence}. A key tension across all these representations is that knowledge accumulation typically proceeds at ingestion time, independent of any future query. Systems must anticipate which facts, relationships, or observations will prove relevant, an inherently lossy process. When precomputed representations do not align with query-time information needs, irrelevant context dilutes the retrieval signal. No existing representation scheme targets only temporally grounded events for structured indexing while retaining raw dialogue for general semantic queries, forcing systems to choose between comprehensive extraction and retrieval simplicity.

\subsection{Retrieval-Augmented Architectures}\label{subsec:retrieval_architectures}

To address context entropy, many system architectures implement some form of retrieval to locate relevant information at query time. Sparse retrieval methods like BM25 excel at exact lexical matching and are computationally efficient, but miss semantic variations and synonymy (\cite{bm25}). Dense retrieval using learned embeddings captures semantic similarity but struggles with precise term matching and rare entities (\cite{DPR}) and introduces computational overhead in generating and storing embeddings. Hybrid approaches that fuse both modalities (typically via reciprocal rank fusion or learned reranking) have emerged as the dominant paradigm in RAG systems, consistently outperforming either method alone (\cite{blendedrag}, \cite{contextualretrieval}). More sophisticated conversational memory systems combine three or more retrieval modalities in parallel, such as cosine similarity, BM25, and graph traversal (\cite{zep}) or pair dense-sparse hybrid search with fine-tuned cross-encoders for reranking (\cite{emergence}). Simpler systems rely on vector-only cosine similarity with configurable top-k retrieval (\cite{mastra}).

Within the retrieval pipeline, pre-retrieval query processing has proven effective for improving recall in document RAG. Query rewriting reformulates user queries to fix ambiguities, resolve pronouns, and better align with indexed content \cite{ma_query_2023}. Hypothetical document embeddings (HyDE) generate synthetic answers to bridge the query-document semantic gap \cite{gao_hyde_2022}. Query decomposition breaks complex multi-hop questions into sequential sub-queries, each retrievable independently \cite{gao_rag_survey_2024}. These techniques are well-established in document retrieval, and recent work has applied query rewriting and decomposition to tool selection \cite{toolshed}. However, no conversational memory system applies query-aware retrieval guidance to long-term memory, where questions span categories (temporal reasoning, knowledge updates, preference recall) that each demand fundamentally different retrieval strategies.

Beyond static retrieval pipelines, agentic RAG introduces a tool-calling paradigm in which an LLM autonomously decides when and how to retrieve additional context, iteratively refining its search until sufficient evidence is gathered (\cite{agenticrag}). This approach transforms retrieval from a single-shot operation into a multi-step reasoning process, as the agent can decompose complex queries, issue multiple targeted searches, and synthesize results across retrieval rounds. Some conversational memory systems deploy tool-calling research agents that dynamically expand context during inference when initial retrieval proves insufficient (\cite{honcho}). Others introduce autonomous retrieval controllers with evidence-gap trackers that route between different retrieval actions based on accumulated evidence (\cite{memr3}). Although agentic retrieval improves recall on complex, multi-hop queries, it introduces latency and cost at inference time, creating a tradeoff between retrieval thoroughness and system responsiveness. Existing retrieval architectures also lack query-conditioned adaptation of the retrieval strategy itself, applying the same pipeline regardless of whether a query demands temporal filtering, semantic matching, or cross-session aggregation.

\subsection{Summarization and Fact Extraction}\label{subsec:summarization_fact_extraction}

Converting raw conversation history into compact, retrievable representations can require either summarization or structured fact extraction. Summarization approaches condense dialogue into natural language summaries at varying granularities: recursive summarization progressively compresses conversation history across sessions while event-based methods like elementary discourse unit (EDU) extraction rewrite sessions into self-contained, event-like statements that normalize entities and resolve coreferences (\cite{recursum}). Fact extraction takes a more structured approach, decomposing conversations into atomic knowledge units such as entity-relation triples, timestamped observations, or tagged assertions. Some architectures generate timestamped natural language observations via monitoring agents (\cite{mastra}). Others derive latent assertions with confidence scores through offline reasoning processes (\cite{honcho}). A further distinction is \textit{when} extraction occurs: most systems extract at ingestion time independent of any query, while query-conditioned variants additionally extract facts from retrieved turns in response to a specific question, improving relevance at the cost of additional inference-time computation (\cite{emergence}). However, no existing system selectively structures only temporal events while preserving raw dialogue for semantic retrieval, leaving a gap in architectures that balance extraction specificity with conversational context preservation.

\section{The Chronos Architecture}\label{sec:method}

Conversational memory systems must balance the overhead of structured knowledge building against the temporal blindness of pure turn-level retrieval. Chronos addresses this by selectively structuring only temporally grounded events while preserving raw dialogue for semantic retrieval (Figure~\ref{fig:final_output}). The system consists of four main components: (1) the event extraction pipeline that identifies timestamped occurrences from conversation text (subsection \ref{subsec:event_extraction}), (2) the dynamic prompting system, extending query rewriting to long-term memory by generating per-question retrieval guidance (subsection \ref{subsec:dynamic_prompting}), (3) an initial retrieval stage using dense search with reranking (subsection \ref{subsec:initial_retrieval}), and (4) the Chronos Agent with tool-calling capabilities for answering memory queries (subsection \ref{subsec:chronos_agent}). Chronos maintains two complementary calendars: an event calendar containing extracted temporal events with structured datetime ranges, and a turn calendar preserving raw conversational exchanges, enabling accurate recall for queries requiring time-grounded reasoning.

\subsection{Event Extraction}\label{subsec:event_extraction}

The event extraction pipeline identifies temporally-grounded occurrences from conversation text using LLM-powered extraction. Given a conversation turn with timestamp $t_{\text{conv}}$, the extractor identifies events if they have all of the following: $\langle \text{subject, verb, object} \rangle$. The pipeline implements multi-resolution temporal normalization to convert natural language time references into precise ISO 8601 datetime ranges. Each event receives both \texttt{start\_datetime} and \texttt{end\_datetime} fields that capture the full temporal span of when the event could have occurred. For ambiguous expressions like "recently" or "last month," the system computes appropriate temporal windows centered on or relative to $t_{\text{conv}}$ with appropriate granularity for the expression. This representation enables precise temporal filtering during retrieval by encoding all possible times the event could have occurred rather than single point estimates.

Beyond structured extraction, the system generates 2-4 lexical aliases for each event using completely different vocabulary to improve text search recall. These aliases paraphrase the event using synonyms, related terms, or category descriptors that avoid repeating key nouns from the original text. For example, "bought Fitbit" generates aliases like "picked up a fitness tracker," "got a step counter," and "purchased a wearable." This alias generation enables robust keyword matching when users query their memory using alternative phrasings.

To extract events, within each session, turns are passed to the extractor in batches of maximum 25 turns, with 5-turn overlap between chunks. However, most sessions contain fewer than 25 turns, thus fitting into a single batch. This event-only representation intentionally avoids global entity graphs or comprehensive fact extraction, structuring only what is necessary for temporal reasoning.

Once extracted, events are embedded using the text-embedding-3-large model and uploaded to the event calendar index, while raw turns are embedded and uploaded to the turn calendar index, enabling independent retrieval over each representation.

\subsection{Dynamic Prompting}\label{subsec:dynamic_prompting}

Chronos introduces dynamic prompting for long-term memory, extending query rewriting from the document retrieval literature \cite{ma_query_2023, gao_rag_survey_2024} to conversational memory. Rather than reformulating the search query itself, dynamic prompting analyzes each question to generate retrieval guidance tailored to the agent's reasoning process. Given a user query $q$, the system generates a custom instruction preamble that directs the agent's attention to relevant information dimensions and retrieval strategies. The template generator uses an LLM meta-prompt to analyze the question structure and produce tailored guidance. The meta-prompt instructs the generator to extract question targets (entities, attributes, temporal constraints, operations) and output 1-5 concrete bullets describing what information the agent should retrieve. For example, given "What camera lens did I buy most recently?", the generator outputs: "Pay close attention to the following information (current and past): Details about camera lens purchases, specifically the most recent purchase and the lens type/model."

The generated preamble integrates into the agent's system prompt alongside retrieval tool descriptions, chain-of-thought guidelines, and pre-retrieved context as detailed in Subsection~\ref{subsec:initial_retrieval}. This approach enables the system to dynamically adjust retrieval breadth, temporal handling, and reasoning patterns without requiring separate models or architectures for different question types. The template generator runs once per question during inference, using a small, efficient model (Gemini 3 Flash). Static prompts cannot anticipate the diversity of long-term memory queries, which range from temporal filtering ("What did I do last March?") to preference recall ("What kind of coffee do I like?") to cross-session aggregation ("How many times did I exercise?"). By generating query-specific guidance, the system adapts its retrieval strategy without requiring separate prompt templates or classifiers for each question category.

\subsection{Initial Retrieval}\label{subsec:initial_retrieval}

Given a user query $q$, the initial retrieval stage provides the agent with conversational context from the turn calendar before tool-based reasoning begins. Retrieval proceeds through a three-stage pipeline: vector search, reranking, and context expansion. First, dense vector search queries the turn calendar to retrieve the top 100 conversation exchanges based on cosine similarity between the query embedding $\mathbf{v}_q$ and turn embeddings $\mathbf{v}_t$. Second, Chronos applies cross-encoder reranking to improve relevance, using Cohere Rerank v3 to rescore the 100 retrieved candidates based on semantic similarity between the query and turn text. After reranking, the top-15 most relevant turns are selected. Finally, each of the 15 selected turns is expanded with conversational context by retrieving one turn before and one turn after from the same conversation session. This context window provides surrounding dialogue that helps the agent understand references, continuity, and conversational flow.

Retrieved turns are formatted into hierarchical natural language context blocks organized by conversation date. Turns group by session with date headers like "Session 1 (2024-02-15)" to make temporal relationships explicit. This pre-retrieved context initializes the agent with relevant background before tool-based reasoning, reducing the need for redundant searches during inference while providing comprehensive coverage of semantically relevant conversations. Finally, the agent is presented with the original question and its date. 

\subsection{Chronos Agent}\label{subsec:chronos_agent}

The Chronos Agent is an LLM agent with native tool-calling capabilities for iterative memory retrieval. The agent receives the fully assembled dynamic prompt (detailed in Subsection~\ref{subsec:dynamic_prompting}) and access to search tools for dynamic re-querying during inference. We equip the agent with vector search tools and grep-based text search tools for both the turn calendar and event calendar. Vector search tools (\texttt{search\_turns}, \texttt{search\_events}) enable semantic retrieval by querying each calendar's index with agent-generated keywords or rephrased queries. Grep tools (\texttt{grep\_turns}, \texttt{grep\_events}) enable exact keyword matching on local files. The grep capability proves particularly valuable when users reference specific entities or exact phrases that vector search may miss due to embedding similarity thresholds.

The agent follows a ReAct reasoning pattern, alternating between thought generation, tool calling, and observation processing \cite{react}. At each step, the agent decides whether to answer directly using pre-retrieved context, invoke search tools to gather additional information, or use grep tools for precise keyword matching. Tool calls execute asynchronously with automatic retry logic for robustness against transient failures. Retrieved results from tool calls are appended to the agent's message history, progressively building up contextual information until the agent can confidently answer the query. The agent is allowed to select the top-k to retrieve, and, for dense retrieval, reranking from k=100 is applied. Notably, reranking is applied to the original question rather than the agent's query. This design allows the agent to iteratively constrain retrieval by datetime range, cross-reference events with source dialogue, and expand the search space when evidence is insufficient, resulting in more reliable long-horizon reasoning.

\subsection{Benchmarks}\label{subsec:evaluation}

We evaluate Chronos on the LongMemEvalS Benchmark. After reasoning and tool-calling, the Chronos Agent outputs a hypothesis. From there, we implement LongMemEval's LLM judge, which compares the hypothesis to the ground truth, routing to a specific prompt based on the question's category. Manual inspection reveals benchmark limitations. Question 6d550036 asks ``How many projects have I led or am currently leading?'' with reference answer 2, yet the history explicitly mentions more than three projects with clear leadership statements. There are also issues with the evaluation methodology. Question 75f70248 asks about sneezing in the living room with a preference rubric requiring mention of the user's cat Luna and HEPA filters. Our system addresses the new cat as a potential source of allergies yet receives an incorrect judgment, highlighting LLM-as-judge variability. Many of these issues have been raised in the repository for the dataset \cite{longmemval_github}.

\section{Experiments}\label{sec:experiments}

We evaluate Chronos against state-of-the-art conversational memory systems including EmergenceMem Internal, Honcho, Mastra, and Zep. To ensure fair comparison, we report results under two configurations: Chronos Low, which uses GPT-4o as the generation model to match the evaluation setup used by all compared systems, and Chronos High, which uses Claude Opus 4.6 to explore performance at a higher model capability tier. Chronos Low achieves 92.60\% accuracy, establishing state-of-the-art performance among practical methods evaluated with GPT-4o, while Chronos High reaches 95.60\%, demonstrating the increased reasoning capabilities of more recent models.

\subsection{Experimental Settings}\label{subsec:experimental_settings}

\paragraph{Baselines.} Chronos Low is compared against four existing practical conversational memory systems evaluated on LongMemEval: EmergenceMem Internal, Honcho, Mastra, and Zep, most of which report results using GPT-4o. For Chronos High, the comparison narrows to the subset of these systems that report results under stronger LLM configurations. Full evaluation results across a diverse set of LLMs can be found in Appendix~\ref{sec:appendix_longmemeval}. All runs use the text-embedding-3-large model with temperature set to 0.

\paragraph{Metrics.} Accuracy serves as the primary metric, computed as the percentage of questions answered correctly across all 500 questions. Where possible, a breakdown of accuracy across each of the six categories is also provided.

\paragraph{Implementation.} All models are accessed via their respective commercial APIs at standard pricing tiers. We set temperature to 0 for all generation calls to ensure deterministic outputs.

\subsection{Results}\label{subsec:main_results}

\begin{figure}[t]
\centering
\includegraphics[width=1\textwidth]{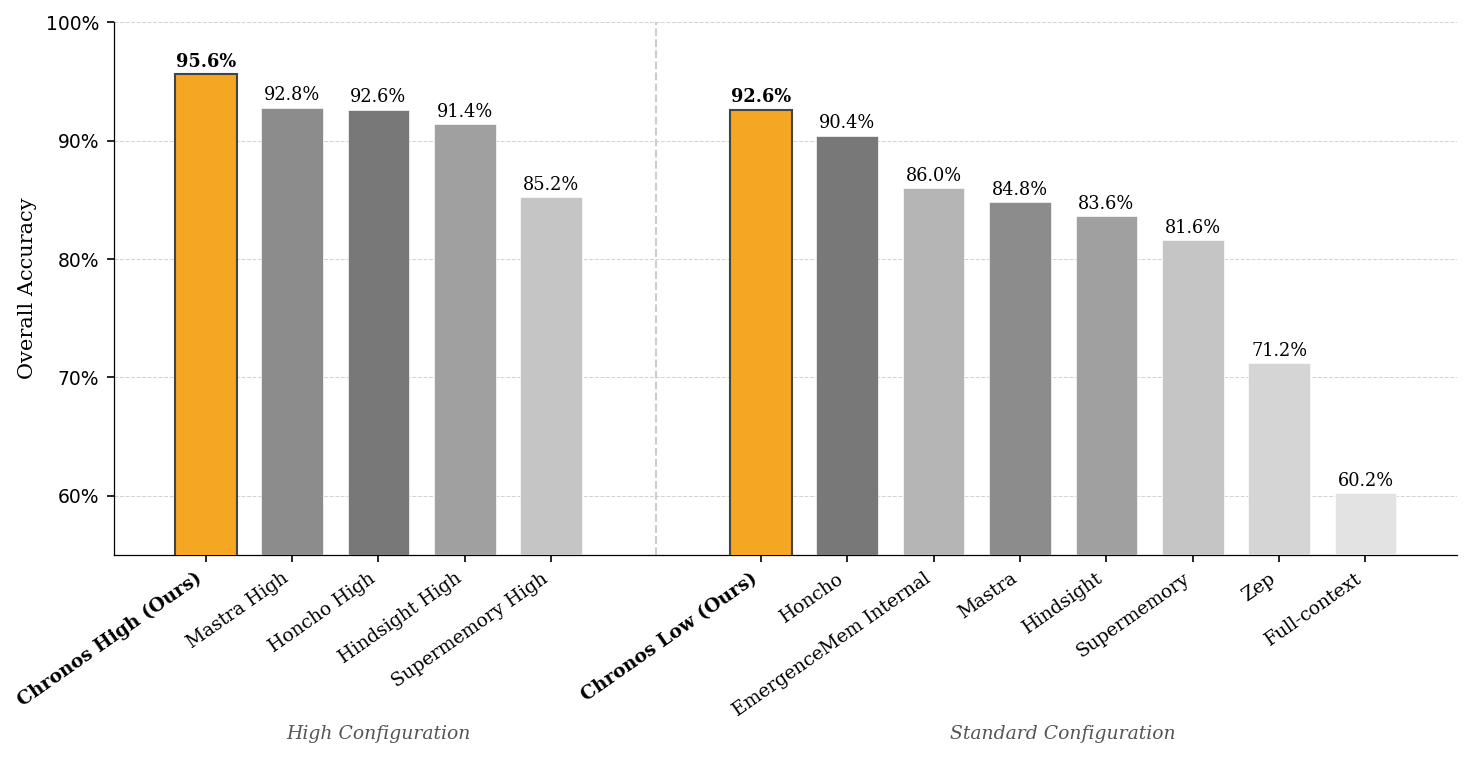}\caption{Overall Benchmark Accuracy on both High and Low Configurations. Note: High configurations refer to evaluations with advanced frontier models, such as Opus 4.6 and Gemini 3 Pro. Standard configurations refer to the traditional evaluated model, GPT-4o, or similar models.}
\label{fig:low_high_comp}
\end{figure}

Table \ref{tab:main_results} presents the overall and category-level accuracy for Chronos compared to baseline conversational memory systems. Chronos achieves 92.60\% overall accuracy, representing the highest performance among non-oracle methods on LongMemEval. Compared to EmergenceMem Internal (86\%), Chronos improves by 7.67\% relative to EmergenceMem Internal overall, with particularly strong gains on multi-session aggregation and knowledge-update tracking. Against Honcho, Chronos improves by 2.43\% relative to Honcho overall while using a weaker model, demonstrating that dual turn-level and event-level retrieval with explicit temporal normalization outperforms event-only approaches. Figure~\ref{fig:low_high_comp} visualizes these performance differences.
\begin{table}[t]
\caption{Comparison of Chronos Low with state-of-the-art conversational memory systems on LongMemEval.}
\label{tab:main_results}
\centering
\begin{tabular}{lcccccccc}
\toprule
\multirow{2}{*}{Method} & \multirow{2}{*}{Overall} & \multicolumn{6}{c}{Category-Level Accuracy (\%)} \\
\cmidrule(lr){3-8}
& & KU & MS & SSA & SSP & SSU & TR \\
\midrule
\textbf{Chronos Low (Ours)} & \textbf{92.60} & \textbf{96.15} & \textbf{91.73} & \textbf{100.00} & 80.00 & 94.29 & \textbf{90.23} \\
Honcho$^{\dagger}$ & 90.40 & 94.87 & 84.96 & 96.43 & \textbf{90.00} & 94.29 & 88.72 \\
EmergenceMem Internal & 86.00 & 83.33 & 81.20 & \textbf{100.00} & 60.00 & \textbf{98.57} & 85.71 \\
Mastra & 84.80 & 85.90 & 79.70 & 82.14 & 73.33 & \textbf{98.57} & 85.71 \\
Supermemory & 81.60 & 88.50 & 71.40 & 96.40 & 70.00 & 97.10 & 76.70 \\
Hindsight$^{\ddagger}$ & 83.60 & 84.60 & 79.70 & 94.60 & 66.70 & 95.70 & 79.70 \\
Zep & 71.20 & 83.30 & 57.90 & 80.40 & 56.70 & 92.90 & 62.40 \\
Full-context  & 60.20 & 78.20 & 44.30 & 94.60 & 20.00 & 81.40 & 45.10 \\
\bottomrule
\end{tabular}

\vspace{0.5em}
\small
$^{\dagger}$Honcho evaluated on Claude Haiku 4.5, not GPT-4o, so results are not directly comparable to other existing systems.\\
$^{\ddagger}$Hindsight evaluated with OSS-20B as the actor model and used OSS-120B as the judge model; results are not directly comparable to GPT-4o-judged systems.
\end{table}
\begin{table}[t]
\captionsetup{justification=centering}
\caption{Comparison of systems with more advanced LLMs on LongMemEval.}
\label{tab:llm_comparison}
\centering
\begin{tabular}{lcccccccc}
\toprule
\multirow{2}{*}{Method} & \multirow{2}{*}{Overall} & \multicolumn{6}{c}{Category-Level Accuracy (\%)} \\
\cmidrule(lr){3-8}
& & KU & MS & SSA & SSP & SSU & TR \\
\midrule
\textbf{Chronos High (Ours)} & \textbf{95.60} & \textbf{100.00} & \textbf{88.72} & \textbf{100.00} & \textbf{100.00} & \textbf{98.57} & \textbf{95.50} \\
Honcho$^{\dagger}$ & 92.60 & - & - & - & - & - & - \\
Mastra & 92.80 & 94.90 & 87.20 & 96.40 & 90.00 & 97.10 & 94.00 \\
Supermemory & 85.20 & 89.70 & 76.70 & 98.20 & 70.00 & 98.60 & 82.00 \\
Hindsight$^{\ddagger}$ & 91.40 & 94.90 & 87.20 & 96.40 & 80.00 & 97.10 & 91.00 \\
\bottomrule
\end{tabular}

\vspace{0.5em}
\small
$^{\dagger}$Category-level accuracy not reported.\\
$^{\ddagger}$Evaluated with an OSS-120B judge model; results are not directly comparable to systems evaluated with the official benchmark judge.
\end{table}

Chronos Low demonstrates strong performance across all six LongMemEvalScategories. The largest gains appear on knowledge-update tracking (KU: 96.15\%), where Chronos outperforms all baselines including Honcho (94.87\%) and EmergenceMem Internal (83.33\%), and on multi-session aggregation (MS: 91.73\%), where it improves over Honcho by 7.97\% relative and over EmergenceMem Internal by 12.97\% relative. Chronos also outperforms EmergenceMem Internal and Mastra on temporal reasoning (TR: 90.23\% vs.\ 85.71\% for both). It achieves perfect accuracy on single-session assistant recall (SSA: 100\%), matching EmergenceMem Internal, and matches Honcho and EmergenceMem Internal on single-session user recall (SSU: 94.29\%). The one category where Chronos trails is single-session preference recall (SSP: 80.00\%), where Honcho achieves 90.00\%. 

In addition, Chronos High achieves the highest reported performance on LongMemEvalxS, with an overall accuracy of 95.60\% (a 3.02\% relative improvement over prior records). Honcho, Hindsight, and Mastra evaluated their systems using Gemini 3 Pro, and Supermemory evaluated their system using GPT-5. Across categories, Chronos High also achieves state-of-the-art performance, scoring 88.72\% on multi-session aggregation, 100\% on single-session assistant, single-session preference, and knowledge-update questions, 98.57\% on single-session user questions, and 95.50\% on temporal reasoning. While this represents a slight regression on multi-session questions from Chronos Low, every other category shows significant improvement. 

\subsection{Discussion}\label{subsec:analysis}

To understand the sources of Chronos's performance improvements, we analyze system behavior across the six LongMemEvalScategories. Multi-session aggregation queries like "How many times did I exercise in May?" require identifying all mentions of a specific activity within a temporal constraint. Chronos's calendar-structured event index enables efficient temporal filtering, retrieving only events within the specified month rather than relying on semantic similarity alone. In addition, strictly extracting time-based events reduces retrieval entropy, ensuring the model is only retrieving events.

Temporal reasoning queries such as "What did I do the week after my vacation?" demand relative date calculation and sequence understanding. Chronos's multi-resolution temporal normalization converts relative expressions like "the week after" into precise date ranges by calculating offsets from reference timestamps. This capability contrasts with turn-level systems that treat temporal expressions as ungrounded strings, forcing the LLM to infer temporal relationships from conversational context alone without structured temporal support. By converting conversational time into executable datetime ranges and exposing them to the agent as retrieval constraints, Chronos shifts temporal reasoning from string interpretation to structured filtering.

For the remaining categories, different Chronos components drive the gains. Knowledge-update queries like "Where do I currently work?" require distinguishing between current and outdated information when the same attribute is mentioned with different values across sessions; Chronos's event extraction identifies each workplace mention as a separate timestamped event, enabling the agent to retrieve all mentions and select the most recent based on datetime ordering. For single-session preference questions, the query rewriting mechanism in dynamic prompting directs the model towards the relevant information dimensions, supporting a deeper understanding of the user's preferences and interests. More broadly, Chronos's agentic design allows it to proactively retrieve additional context when initial evidence is insufficient. Note that Honcho's improvement over Chronos Low on SSP reflects its use of a more powerful model (Claude Haiku 4.5) rather than an architectural advantage.

\begin{figure}[t]
\centering
\includegraphics[width=0.8\textwidth]{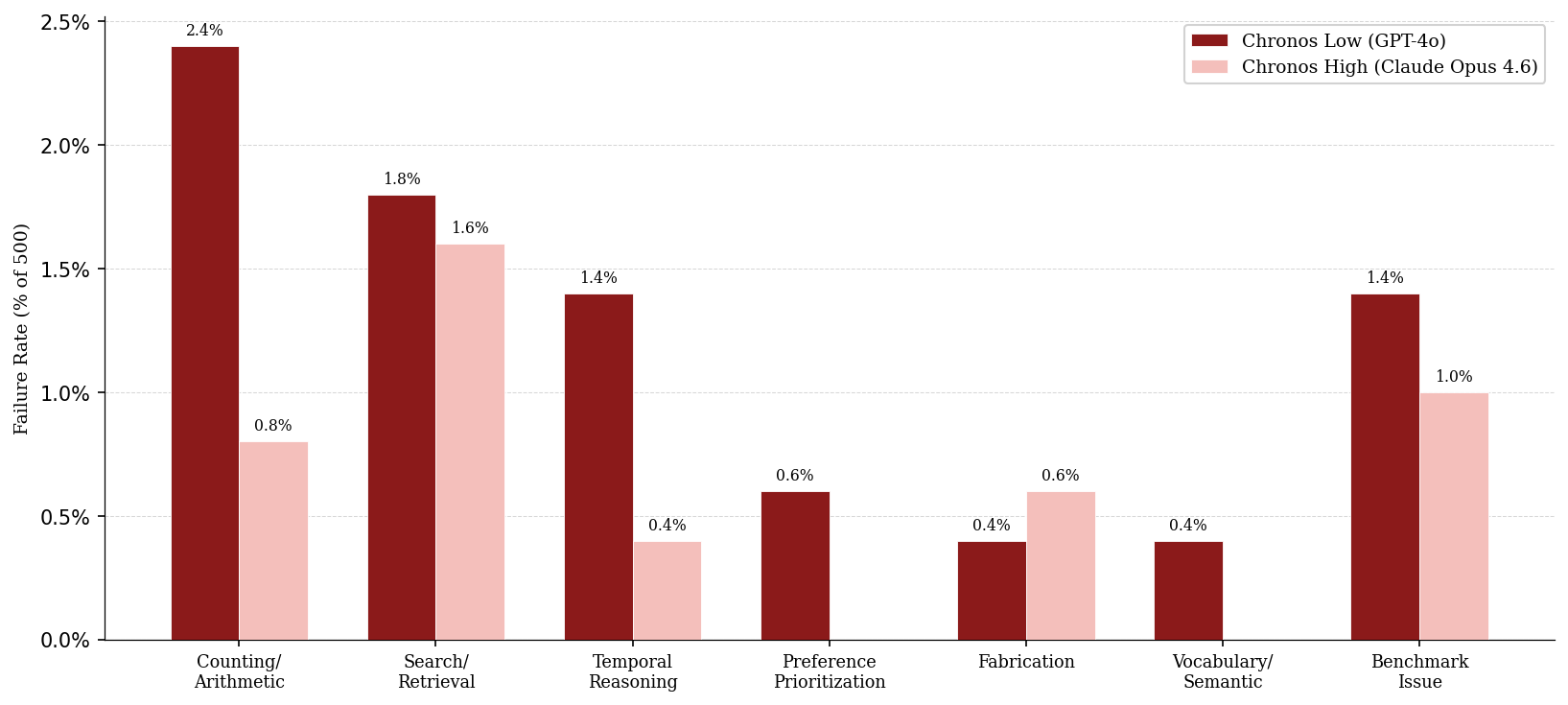}
\caption{Category-wise Error rate comparison across Chronos variations. Chronos High demonstrates marked improvement on nearly every category.}
\label{fig:error_comparison}
\end{figure}

Figure~\ref{fig:error_comparison} shows the limitations of GPT-4o as a reasoning model. When moving from Chronos Low to Chronos High, we see a halving in the counting and arithmetic errors and significant reductions in most other categories. Notably, retrieval failures remain the most common error category for both models, showing that even with specific guidance, LLMs still fail to retrieve over large amounts of data. In addition, there is no performance increase regarding fabrication. These errors frequently occurred with abstention questions, where the model made assumptions about the user instead of refusing to answer. Additional errors appear attributable to the benchmark's ground truth answers, as discussed in Section~\ref{subsec:evaluation}. The difference in performance between Chronos Low and Chronos High on this category can be explained by open-endedness in some of the benchmark's questions.

\subsection{Ablation Studies}\label{subsec:ablation}

\begin{table}[t]
\centering
\small
\setlength{\tabcolsep}{4pt}
\begin{tabular}{llrrrrrrr}
\toprule
Model & Ablation & Overall & KU & MS & SSA & SSP & SSU & TR \\
\midrule
\multirow{8}{*}{Chronos High}
& Chronos    & \textbf{94.8} & \textbf{100.0} & \textbf{90.3} & \textbf{100.0} & \textbf{100.0} & \textbf{93.8} & 93.5 \\
& No Initial Retrieval   & 91.4 & 100.0 & 80.6 & 100.0 & 85.7 & 93.8 & 93.5 \\
& No Dynamic Prompting   & 94.8 & 100.0 & 87.1 & 100.0 & 100.0 & 93.8 & \textbf{96.8} \\
& No Rerank              & 92.2 & 100.0 & 87.1 & 100.0 & 85.7 & 87.5 & 93.5 \\
& No Date Filter         & 92.2 & 94.4 & 83.9 & 100.0 & 85.7 & 93.8 & 96.8 \\
& Grep Only (no vector)  & 87.1 & 94.4 & 83.9 & 61.5 & 100.0 & 93.8 & 90.3 \\
& Vector Only (no grep)  & 83.6 & 100.0 & 80.6 & 100.0 & 85.7 & 62.5 & 80.6 \\
& Turns Only (no events) & 92.2 & 94.4 & 87.1 & 100.0 & 85.7 & 87.5 & 96.8 \\
\midrule
\multirow{8}{*}{Chronos Low}
& Chronos    & \textbf{93.1} & 94.4 & \textbf{90.3} & \textbf{100.0} & 85.7 & 87.5 & \textbf{96.8} \\
& No Initial Retrieval   & 76.7 & 83.3 & 64.5 & 100.0 & 71.4 & 81.2 & 74.2 \\
& No Dynamic Prompting   & 78.4 & 83.3 & 77.4 & 100.0 & 42.9 & \textbf{93.8} & 67.7 \\
& No Rerank              & 81.0 & 77.8 & 74.2 & 100.0 & 85.7 & 87.5 & 77.4 \\
& No Date Filter         & 78.4 & 88.9 & 71.0 & 100.0 & 57.1 & 68.8 & 80.6 \\
& Grep Only (no vector)  & 77.6 & \textbf{100.0} & 64.5 & 53.8 & \textbf{100.0} & 87.5 & 77.4 \\
& Vector Only (no grep)  & 80.2 & 94.4 & 80.6 & 100.0 & 85.7 & 62.5 & 71.0 \\
& Turns Only (no events) & 58.6 & 55.6 & 51.6 & 100.0 & 42.9 & 43.8 & 61.3 \\
\bottomrule
\end{tabular}
\caption{Ablation results grouped by model configuration on a 116-question sample of the benchmark.}
\label{ablation}
\end{table}

To understand the contribution of each component in Chronos, we conduct a series of ablation studies by systematically removing one component at a time and evaluating on a stratified 116-question subset of our benchmark. We chose a sample of the benchmark due to the large number of ablations we chose to apply. We present the overall and per-category results for these ablations in Table~\ref{ablation}. For Chronos Low, each component is essential to maintaining state-of-the-art performance. Without pre-reasoning instruction (dynamic prompting and initial retrieval), performance drops by around 14 to 16 points. Restricting search methods and removing reranking also caused steep drops in performance. The steepest drop, however, was caused by removing access to the events index, which almost halved accuracy.

The larger performance drops observed for Chronos Low under each ablation condition reflect the degree to which each Chronos component compensates for reduced backbone capability. Chronos Low uses a comparatively older and less capable model, so it cannot recover missing signals independently. Removing event context, for instance, reduces Chronos Low's performance by 34.5 points compared to only 2.6 for Chronos High. However, the ablation studies further demonstrate that Chronos still scales with model capability: while the performance increase is more modest for Chronos High, most components contribute meaningfully, confirming that the unique enhancements of the Chronos architecture deliver consistent gains regardless of the underlying model's capability. Interestingly, under the high configuration, dynamic prompting does not affect performance at all, reflecting more advanced models' capabilities to distinguish between query types without advanced guidance. 

\section{Limitations}\label{sec:limitations}

We identify three primary limitations of our approach. First, Chronos's dual indexing architecture requires maintaining separate vector indexes for conversation turns and extracted events, increasing storage requirements compared to turn-only systems. We address this through efficient sparse event extraction that focuses only on temporally-grounded state transitions rather than comprehensive fact extraction, keeping the event index compact. Second, event extraction using LLM-powered processing adds offline computation costs during indexing, similar to all existing systems. We mitigate this through batched processing with overlapping chunks (25 turns per batch, 5-turn overlap) and use of efficient small models for extraction. Third, parallel retrieval operations over both turns and events at query time increase inference complexity. We optimize this through pre-retrieval of turn context before agent reasoning begins, reducing redundant searches during inference. Future work on more efficient memory operations and shared event histories, as discussed in Section~\ref{sec:conclusion}, could further alleviate these constraints.

\section{Conclusion}\label{sec:conclusion}

As conversational AI systems scale to extended multi-turn interactions spanning weeks or months, existing retrieval-augmented generation approaches for conversational memory fail to accurately handle temporal queries requiring precise time-grounded reasoning, such as "What did I do the week after my vacation?" or "When did I start that medication?" In this work, we introduce Chronos, a date-aware memory framework that maintains dual calendars: an event calendar of extracted temporal events with structured datetime ranges, and a turn calendar preserving raw conversational exchanges. Chronos combines these representations with turn-level dense retrieval, text search, and query-conditioned dynamic prompting, enabling agents to answer temporally grounded queries about past conversations with high accuracy. Chronos extracts and indexes timestamped events from conversations using LLM-powered extraction with multi-resolution time handling: Chronos preserves explicit dates exactly, calculates relative references from conversation timestamps, and resolves ambiguous temporal expressions to appropriate time ranges. We evaluate Chronos across multiple state-of-the-art LLMs on the LongMemEvalSbenchmark comprising 500 questions spanning six categories of conversational memory tasks, including knowledge-update tracking, multi-session aggregation, single-session recall, and temporal reasoning. With GPT-4o, Chronos achieves 92.60\% accuracy, surpassing prior systems by 7.67\% relative and establishing state-of-the-art performance among practical (non-oracle) conversational memory systems. Our approach demonstrates particularly strong performance on multi-session aggregation (91.73\%) and temporal reasoning (90.23\%), categories where prior methods struggle to maintain accuracy. In addition, Chronos High achieves 95.60\% on the benchmark, improving over prior records by 3.02\% relative and representing the highest reported score. These results show that integrating structured temporal representations with turn-level retrieval enables accurate time-grounded memory without requiring oracle access to ground truth conversation histories. More broadly, Chronos demonstrates that persistent conversational agents do not require comprehensive knowledge graph construction to achieve high-fidelity long-horizon memory. Structuring fine-grained temporal events while preserving full dialogue context is sufficient to support accurate update tracking, preference recall, and relative date reasoning across months of interaction. Looking forward, an important direction is enabling models to not only retrieve structured events but also learn from them, for example by updating model weights using accumulated event traces. Another promising avenue is improving the scalability of such systems, including more efficient memory operations and shared event histories that support multi-agent or multi-user interactions.

\clearpage

\bibliography{colm2026_conference}

@misc{rag_survey,
      title={From Human Memory to AI Memory: A Survey on Memory Mechanisms in the Era of LLMs}, 
      author={Yaxiong Wu and Sheng Liang and Chen Zhang and Yichao Wang and Yongyue Zhang and Huifeng Guo and Ruiming Tang and Yong Liu},
      year={2025},
      eprint={2504.15965},
      archivePrefix={arXiv},
      primaryClass={cs.IR},
      url={https://arxiv.org/abs/2504.15965}, 
}

@misc{honcho,
  title={Benchmarking Honcho},
  author={McCormick, Ben and Leer, Courtland},
  year={2025},
  month={December},
  howpublished={\url{https://blog.plasticlabs.ai/research/Benchmarking-Honcho}},
  note={Accessed: 2026-02-22}
}

@misc{memgpt,
      title={MemGPT: Towards LLMs as Operating Systems}, 
      author={Charles Packer and Sarah Wooders and Kevin Lin and Vivian Fang and Shishir G. Patil and Ion Stoica and Joseph E. Gonzalez},
      year={2024},
      eprint={2310.08560},
      archivePrefix={arXiv},
      primaryClass={cs.AI},
      url={https://arxiv.org/abs/2310.08560}, 
}

@misc{longmemeval,
      title={LongMemEval: Benchmarking Chat Assistants on Long-Term Interactive Memory}, 
      author={Di Wu and Hongwei Wang and Wenhao Yu and Yuwei Zhang and Kai-Wei Chang and Dong Yu},
      year={2025},
      eprint={2410.10813},
      archivePrefix={arXiv},
      primaryClass={cs.CL},
      url={https://arxiv.org/abs/2410.10813}, 
}

@misc{locomo,
      title={Evaluating Very Long-Term Conversational Memory of LLM Agents}, 
      author={Adyasha Maharana and Dong-Ho Lee and Sergey Tulyakov and Mohit Bansal and Francesco Barbieri and Yuwei Fang},
      year={2024},
      eprint={2402.17753},
      archivePrefix={arXiv},
      primaryClass={cs.CL},
      url={https://arxiv.org/abs/2402.17753}, 
}

@misc{locomo_bad,
  title={Lies, Damn Lies, \& Statistics: Is Mem0 Really {SOTA} in Agent Memory?},
  author={Chalef, Daniel and Rasmussen, Preston},
  year={2025},
  month={May},
  howpublished={\url{https://blog.getzep.com/lies-damn-lies-statistics-is-mem0-really-sota-in-agent-memory/}},
  note={Accessed: 2026-02-22}
}

@misc{zep,
      title={Zep: A Temporal Knowledge Graph Architecture for Agent Memory}, 
      author={Preston Rasmussen and Pavlo Paliychuk and Travis Beauvais and Jack Ryan and Daniel Chalef},
      year={2025},
      eprint={2501.13956},
      archivePrefix={arXiv},
      primaryClass={cs.CL},
      url={https://arxiv.org/abs/2501.13956}, 
}

@misc{local_to_global,
      title={From Local to Global: A Graph RAG Approach to Query-Focused Summarization}, 
      author={Darren Edge and Ha Trinh and Newman Cheng and Joshua Bradley and Alex Chao and Apurva Mody and Steven Truitt and Dasha Metropolitansky and Robert Osazuwa Ness and Jonathan Larson},
      year={2025},
      eprint={2404.16130},
      archivePrefix={arXiv},
      primaryClass={cs.CL},
      url={https://arxiv.org/abs/2404.16130}, 
}

@misc{emergence,
  title={{SOTA} on {LongMemEval} with {RAG}},
  author={Haley, Paul and Pickett, Marc and Hartman, Jeremy and Dixit, Prakhar},
  year={2025},
  month={June},
  howpublished={\url{https://www.emergence.ai/blog/sota-on-longmemeval-with-rag}},
  note={Accessed: 2026-02-22}
}

@article{bm25,
author = {Robertson, Stephen and Zaragoza, Hugo},
title = {The Probabilistic Relevance Framework: BM25 and Beyond},
year = {2009},
issue_date = {April 2009},
publisher = {Now Publishers Inc.},
address = {Hanover, MA, USA},
volume = {3},
number = {4},
issn = {1554-0669},
url = {https://doi.org/10.1561/1500000019},
doi = {10.1561/1500000019},
journal = {Found. Trends Inf. Retr.},
month = apr,
pages = {333–389},
numpages = {57}
}

@misc{DPR,
      title={Dense Passage Retrieval for Open-Domain Question Answering}, 
      author={Vladimir Karpukhin and Barlas Oğuz and Sewon Min and Patrick Lewis and Ledell Wu and Sergey Edunov and Danqi Chen and Wen-tau Yih},
      year={2020},
      eprint={2004.04906},
      archivePrefix={arXiv},
      primaryClass={cs.CL},
      url={https://arxiv.org/abs/2004.04906}, 
}

@inproceedings{blendedrag,
   title={Blended RAG: Improving RAG (Retriever-Augmented Generation) Accuracy with Semantic Search and Hybrid Query-Based Retrievers},
   url={http://dx.doi.org/10.1109/MIPR62202.2024.00031},
   DOI={10.1109/mipr62202.2024.00031},
   booktitle={2024 IEEE 7th International Conference on Multimedia Information Processing and Retrieval (MIPR)},
   publisher={IEEE},
   author={Sawarkar, Kunal and Mangal, Abhilasha and Solanki, Shivam Raj},
   year={2024},
   month=aug, pages={155–161} }

@misc{contextualretrieval,
      title={Learning Contextual Retrieval for Robust Conversational Search}, 
      author={Seunghan Yang and Juntae Lee and Jihwan Bang and Kyuhong Shim and Minsoo Kim and Simyung Chang},
      year={2025},
      eprint={2509.19700},
      archivePrefix={arXiv},
      primaryClass={cs.IR},
      url={https://arxiv.org/abs/2509.19700}, 
}

@misc{mastra,
  title={Observational Memory: 95\% on {LongMemEval}},
  author={Barnes, Tyler},
  year={2026},
  month={February},
  howpublished={\url{https://mastra.ai/research/observational-memory}},
  note={Accessed: 2026-02-22}
}

@misc{agenticrag,
      title={Self-RAG: Learning to Retrieve, Generate, and Critique through Self-Reflection}, 
      author={Akari Asai and Zeqiu Wu and Yizhong Wang and Avirup Sil and Hannaneh Hajishirzi},
      year={2023},
      eprint={2310.11511},
      archivePrefix={arXiv},
      primaryClass={cs.CL},
      url={https://arxiv.org/abs/2310.11511}, 
}

@misc{memr3,
      title={MemR$^3$: Memory Retrieval via Reflective Reasoning for LLM Agents}, 
      author={Xingbo Du and Loka Li and Duzhen Zhang and Le Song},
      year={2025},
      eprint={2512.20237},
      archivePrefix={arXiv},
      primaryClass={cs.AI},
      url={https://arxiv.org/abs/2512.20237}, 
}

@misc{recursum,
      title={Recursively Summarizing Enables Long-Term Dialogue Memory in Large Language Models}, 
      author={Qingyue Wang and Yanhe Fu and Yanan Cao and Shuai Wang and Zhiliang Tian and Liang Ding},
      year={2025},
      eprint={2308.15022},
      archivePrefix={arXiv},
      primaryClass={cs.CL},
      url={https://arxiv.org/abs/2308.15022}, 
}

@inproceedings{longmemval_github,
  title     = {LongMemEval: Benchmarking Chat Assistants on Long-Term Interactive Memory},
  author    = {Di Wu and Hongwei Wang and Wenhao Yu and Yuwei Zhang and Kai-Wei Chang and Dong Yu},
  booktitle = {Proc.\ of the International Conference on Learning Representations (ICLR)},
  year      = {2025},
  url       = {https://github.com/xiaowu0162/LongMemEval},
  note      = {Benchmark and code available at GitHub},
  archivePrefix = {arXiv},
  eprint    = {2410.10813},
  primaryClass = {cs.CL}
}

@misc{react,
      title={ReAct: Synergizing Reasoning and Acting in Language Models}, 
      author={Shunyu Yao and Jeffrey Zhao and Dian Yu and Nan Du and Izhak Shafran and Karthik Narasimhan and Yuan Cao},
      year={2023},
      eprint={2210.03629},
      archivePrefix={arXiv},
      primaryClass={cs.CL},
      url={https://arxiv.org/abs/2210.03629}, 
}

@online{llm_memory,
  author       = {Jenny Tan},
  title        = {Understanding Large Language Model (LLM) Short-Term and Long-Term Memory},
  year         = {2025},
  month        = {September},
  day          = {24},
  url          = {https://medium.com/@jennytan5522/understanding-large-language-model-llm-short-term-and-long-term-memory-fa1e2d56fc2b},
  note         = {Medium blog post},
}

@misc{ma_query_2023,
      title={Query Rewriting for Retrieval-Augmented Large Language Models},
      author={Xinbei Ma and Yeyun Gong and Pengcheng He and Hai Zhao and Nan Duan},
      year={2023},
      eprint={2305.14283},
      archivePrefix={arXiv},
      primaryClass={cs.CL},
      url={https://arxiv.org/abs/2305.14283},
}

@misc{gao_hyde_2022,
      title={Precise Zero-Shot Dense Retrieval without Relevance Labels},
      author={Luyu Gao and Xueguang Ma and Jimmy Lin and Jamie Callan},
      year={2022},
      eprint={2212.10496},
      archivePrefix={arXiv},
      primaryClass={cs.IR},
      url={https://arxiv.org/abs/2212.10496},
}

@misc{gao_rag_survey_2024,
      title={Retrieval-Augmented Generation for Large Language Models: A Survey},
      author={Yunfan Gao and Yun Xiong and Xinyu Gao and Kangxiang Jia and Jinliu Pan and Yuxi Bi and Yi Dai and Jiawei Sun and Qianyu Guo and Meng Wang and Haofen Wang},
      year={2024},
      eprint={2312.10997},
      archivePrefix={arXiv},
      primaryClass={cs.CL},
      url={https://arxiv.org/abs/2312.10997},
}

@inproceedings{toolshed,
      title={Toolshed: Advanced {RAG}-Tool Fusion for Scalable Real-World Agent Tool Selection},
      author={Elias Lumer and Sahil Sen},
      booktitle={Proc.\ of the International Conference on Agents and Artificial Intelligence (ICAART)},
      year={2025},
}
\bibliographystyle{colm2026_conference}

\clearpage
\onecolumn
\appendix

\section{Additional LongMemEvalSResults}
\label{sec:appendix_longmemeval}

\captionsetup{type=table}
\caption{Model comparison on LongMemEvalSacross 500 total questions. Numbers in parentheses indicate total questions per category. Bold indicates best or tied-best performance in each category.}
\label{tab:longmemeval_appendix}
\centering
\begin{tabular}{lccccccc}
\toprule
\multirow{2}{*}{Model}
& \multirow{2}{*}{Overall (500)}
& \multicolumn{6}{c}{Category-Level Accuracy (\%)} \\ 
\cmidrule(lr){3-8}
& & KU (78) & MS (133) & SSA (56) & SSP (30) & SSU (70) & TR (133) \\
\midrule
Claude Opus 4.6
& \textbf{95.60}
& \textbf{100.00}
& 88.72
& \textbf{100.00}
& \textbf{100.00}
& \textbf{98.57}
& 95.49 \\

GPT-5-mini
& 94.20
& 96.15
& 85.71
& 98.21
& \textbf{100.00}
& 97.14
& \textbf{96.99} \\

Claude Sonnet 4.5
& 94.20
& \textbf{98.72}
& 89.47
& \textbf{100.00}
& 93.33
& \textbf{98.57}
& 91.73 \\

Claude Haiku 4.5
& 94.00
& 96.15
& 88.72
& \textbf{100.00}
& 90.00
& 97.14
& 94.74 \\

GPT-5.2
& 93.80
& 97.44
& 83.46
& \textbf{100.00}
& \textbf{100.00}
& \textbf{98.57}
& 95.49 \\

GPT-4o
& 92.60
& 96.15
& \textbf{91.73}
& \textbf{100.00}
& 80.00
& 94.29
& 90.23 \\

Claude Code Sonnet
& 88.60
& 93.59
& 79.70
& 94.64
& 80.00
& 94.29
& 90.98 \\
\bottomrule
\end{tabular}
\end{document}